\documentclass[runningheads]{llncs}
\usepackage{graphicx}

\usepackage{tikz}
\usepackage{comment}
\usepackage{amsmath,amssymb} 
\usepackage{color}

\usepackage{algorithm}
\usepackage{algpseudocode}

\begin{document}
\pagestyle{headings}
\mainmatter

\title{Data-Efficient Deep Learning Method for Image Classification Using Data Augmentation, Focal Cosine Loss, and Ensemble} 

\titlerunning{Data-Efficient Deep Learning Method for Image Classification}
%
\author{Byeongjo Kim \and
Chanran Kim \and 
Jaehoon Lee \and
Jein Song \and 
Gyoungsoo Park}

\authorrunning{B. Kim et al.}
%

\institute{Zuminternet, Seoul, South Korea \\
\email{\{byeongjo.kim,chanranhari,ljh0128,aimaster,gspark\}@zuminternet.com}}
\maketitle

\begin{abstract}
In general, sufficient data is essential for the better performance and generalization of deep-learning models. However, lots of limitations(cost, resources, etc.) of data collection leads to lack of enough data in most of the areas. In addition, various domains of each data sources and licenses also lead to difficulties in collection of sufficient data. This situation makes us hard to utilize not only the pre-trained model, but also the external knowledge. Therefore, it is important to leverage small dataset effectively for achieving the better performance. We applied some techniques in three aspects: data, loss function, and prediction to enable training from scratch with less data. With these methods, we obtain high accuracy by leveraging ImageNet data which consist of only 50 images per class. Furthermore, our model is ranked 4th in Visual Inductive Printers for Data-Effective Computer Vision Challenge.
\keywords{Data-Efficient, Deep-Learning, Image Classification, Data Augmentation, Cosine Loss, Focal Loss, Ensemble}
\end{abstract}

\section{Introduction}
There are limits to collecting a lot of data used to train deep-learning models. A transfer learning and a pre-training technique trained with large image dataset are used to achieve good performance even if there is small dataset. However, most large image dataset consist of images collected from web, which may be licensed or commercially prohibited~\cite{barz2020deep}. Therefore, in some cases, pre-trained weights trained with ImageNet~\cite{deng2009imagenet} may not be applied for commercial applications.

Nowadays it is known that self-training can replace pre-trained model and even perform better~\cite{zoph2020rethinking,he2019rethinking}. However, self-training requires a lot of data that is not labeled. So data problems eventually arise again. Therefore, even if given dataset is small, we should use that data very effectively.

Visual Inductive Printers for Data-Effective Computer Vision Challenge~\cite{vip2020} is a challenge improving performance using small dataset productively. There are only 50 images per class, and as with the ImageNet dataset, there are total 1000 classes. This challenge also restricts the use of external data such as pre-trained weights, ontology knowledge. Due to the small dataset, it is difficult to perform well with general training.

We solve this problem by experimenting with various techniques. First, we propose LSB Swap data augmentation technique not only to make various data available but also to improve the robustness of the model. Then, we apply Focal Cosine Loss function to enable model training well with even a small amount of data. Last, we apply voting ensemble to compensate for the wrong answers. Through the model applied with these techniques, the Visual Inductive Printers for Data-Effective Computer Vision Challenge recorded 0.67 top-1 accuracy and ranked 4th.

\section{Method}
\subsection{LSB Swap}

\begin{figure}
\centering
\includegraphics[width=\linewidth]{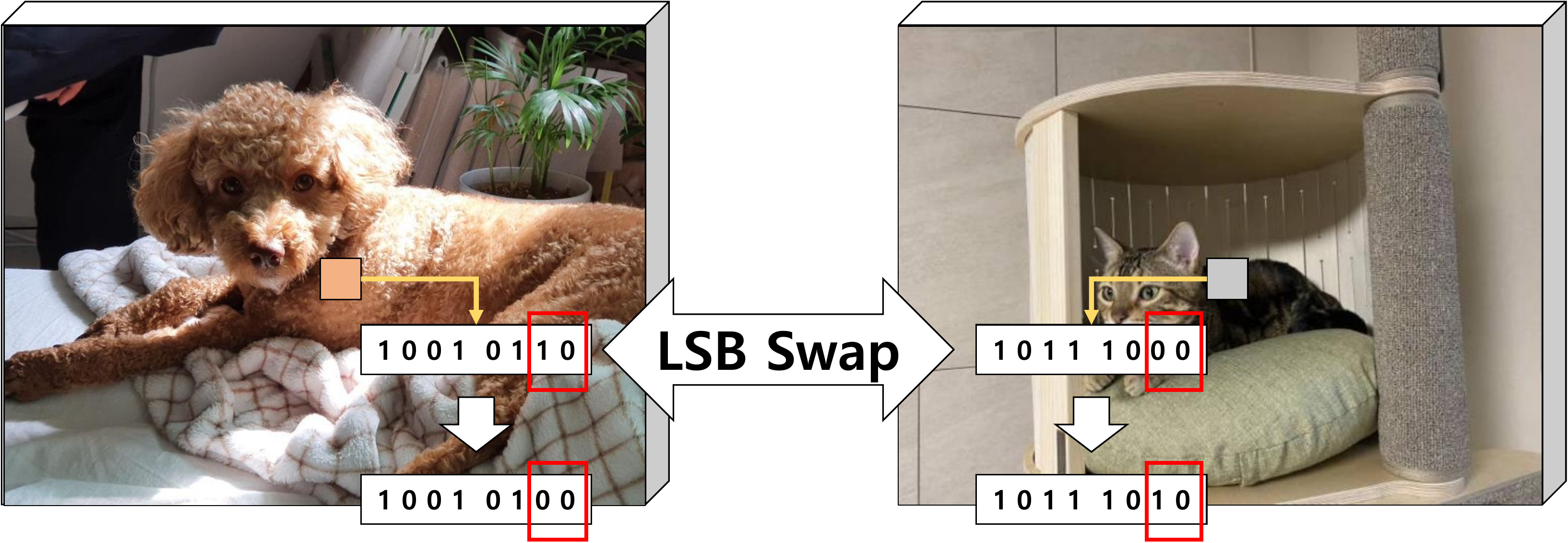}
\caption{Exchange $k$ LSB from all pixels in the image with different images. $k$ is set to 2 in the example.}
\label{fig:lsb}
\end{figure}

We apply two data augmentation techniques in the two aspects. First, RandAugment~\cite{cubuk2020randaugment} is used to transform the data and to create the effect of training with a lot of data. Secondly, LSB(Least Significant Bit) Swap is proposed and applied to focus on structural features for classifying objects. The process of LSB Swap can be seen in Figure~\ref{fig:lsb}.

Because changing a few last bits do not make a difference to be judged by humans, this technique allows data augmentation by swapping non-critical parts of the image classification task. In addition, unlike other data augmentations which use random noise, the LSB Swap helps to use natural noise values present in the images. In environments with less training data, LSB Swap keeps these values from training as much as possible, and helps the model focus on structural features.

\subsection{Focal Cosine Loss}
We used cosine loss, which is known as performing well when training without pre-trained weights and with small dataset~\cite{barz2020deep}. Let $x$ be image, and $y$ be the class label of $x$. $\varphi(y)$ is a one-hot vector mapped each class label. $f_{\theta}(x)$ denotes embedding(i.e., CNN) with learned parameters $\theta$ from $x$ to feature space. $\psi(\cdot)$ denotes embedding(i.e., fully-connected layer) from features into prediction space. The cosine loss is defined as

\begin{align}
L_{cos}(x,y) = 1 - <\varphi(y), \psi(f_{\theta}(x))>
\end{align}

where $<\cdot, \cdot>$ denotes the dot product. For classification accuracy, Barz et al. \cite{barz2020deep} used cosine loss with a categorical cross-entropy loss. In addition, we add a focal loss~\cite{lin2017focal} to focus more on classes which are difficult to classify. The Focal Cosine Loss is defined as


$$FL(p, \hat{p}) = - \sum_{h}p^{(h)}(1-\hat{p}^{(h)})^{\gamma}log(\hat{p}^{(h)})$$

\begin{align}
L_{cos+focal}(x,y) = 1 - <\varphi(y), \psi(f_{\theta}(x))> - \lambda \cdot FL(\varphi(y), g_{\theta}(\psi(f_{\theta}(x))))
\end{align}

where $p^{(h)}$ and $\hat{p}^{(h)}$, respectively, refer to the probability of label $h$ in correct answers and the predictions. Moreover, $g_{\theta}$ denotes an additional fully-connected layer with softmax activation function. Furthermore, we use two hyper-parameters $\gamma$ for focusing and $\lambda$ for combination, following \cite{barz2020deep,lin2017focal}.

\subsection{Plurality Voting Ensemble}
We use plurality voting technique of hard voting ensemble to compensate for the difficult problems through various models and to produce good performance. The plurality voting ensemble algorithm is shown in Algorithm 1. If the predicted confidence score($s_j$) for a particular problem in the base model($B$) is less than a certain $threshold$, it is replaced with voting result($Mode(V_j)$) of predictions for various models($K$). 

\newcommand{\factorial}{\ensuremath{\mbox{\sc Factorial}}}
\begin{algorithm}
\caption{Plurality Voting Ensemble}\label{plurality}
\begin{algorithmic}[1]
\Procedure{PluralityVoting}{$B,K,T$}

\State $B$: Base classification model
\State $K$: $m$ Classification models\Comment{$B$ also belongs to $K$}
\State $T$: $n$ Test data
\State $V_{j}$: Votes from $m$ classifiers using $j$ th test data
\State $O$: results of plurality voting

\For{$j \gets 1$ to $n$}
    \State $s_{j} \gets$ Prediction result of $T_{j}$ from $B$
    
    \If{confidence score of $s_{j}$ $<$ $threshold$}
        \For{$i \gets 1$ to $m$}
        \State $v_{i} \gets$ Prediction result of $T_{j}$ from $K_{i}$
        \State append $v_{i}$ to $V_{j}$
        \EndFor
        \State $O_{j} \gets Mode(V_{j})$
    
    \Else
        \State $O_{j} \gets s_{j}$
        
    \EndIf
\EndFor

\State \Return $O$

\EndProcedure
\end{algorithmic}
\end{algorithm}

\subsection{Other Techniques}
Various techniques are also applied. First, EMA(Exponential Moving Average) model is used to adjust parameters when the model is training, and those parameters are used for testing. Second, drop out and drop connection are used for preventing overfitting. Last, the TTA(Test Time Augmentation) technique is used, which also have a single ensemble effect.

\section{Experiments}
\subsection{Setup}
We use open source named timm~\cite{timm} for training. The code is developed by PyTorch and is capable of applying the recent image classification models and many techniques. All hyper-parameters such as optimizer, learning rate, etc. are set as recommended by timm. For LSB Swap, Focal Cosine Loss, and plurality voting ensemble described above, we developed using Python, NumPy, and PyTorch. The hyper-parameters in Focal Cosine Loss, $\lambda$ and $\gamma$ are set to 0.1 and 2.0 respectively. In addition, $threshold$ for plurality voting ensemble is set to 0.7. When we apply TTA technique, we use the same RandAugment used in training and soft voting ensemble with confidence scores of 1000 classes. Last, k in LSB Swap is set to 2. We trained model using only the data given, not the pre-trained weights and external data. The given training data and validation data are combined to be used in training.

\subsection{Results}

\setlength{\tabcolsep}{4pt}
\begin{table}
\begin{center}
\caption{10 models trained using proposed techniques and given data.}
\label{table:1}
\begin{tabular}{l}
\hline\noalign{\smallskip}
Models\\
\noalign{\smallskip}
\hline
\noalign{\smallskip}
EfficientNet-B3~\cite{tan2019efficientnet}\\
EfficientNet-B3 with TTA\\
EfficientNet-B3 with LSB Swap\\
EfficientNet-B3 with LSB Swap + TTA\\
EfficientNet-B4\\
EfficientNet-B4 with TTA\\
EfficientNet-B5\\
EfficientNet-B5 with TTA\\
EfficientNet-B7\\
EfficientNet-B7 with TTA\\
\hline
\end{tabular}
\end{center}
\end{table}
\setlength{\tabcolsep}{1.4pt}

Table~\ref{table:1} shows that 10 models which is trained with given data and techniques. The RandAugment, cosine loss, and other techniques are applied to all models, and LSB Swap is applied only to EfficientNet-B3.

\begin{figure}
\centering
\includegraphics[width=0.7\linewidth]{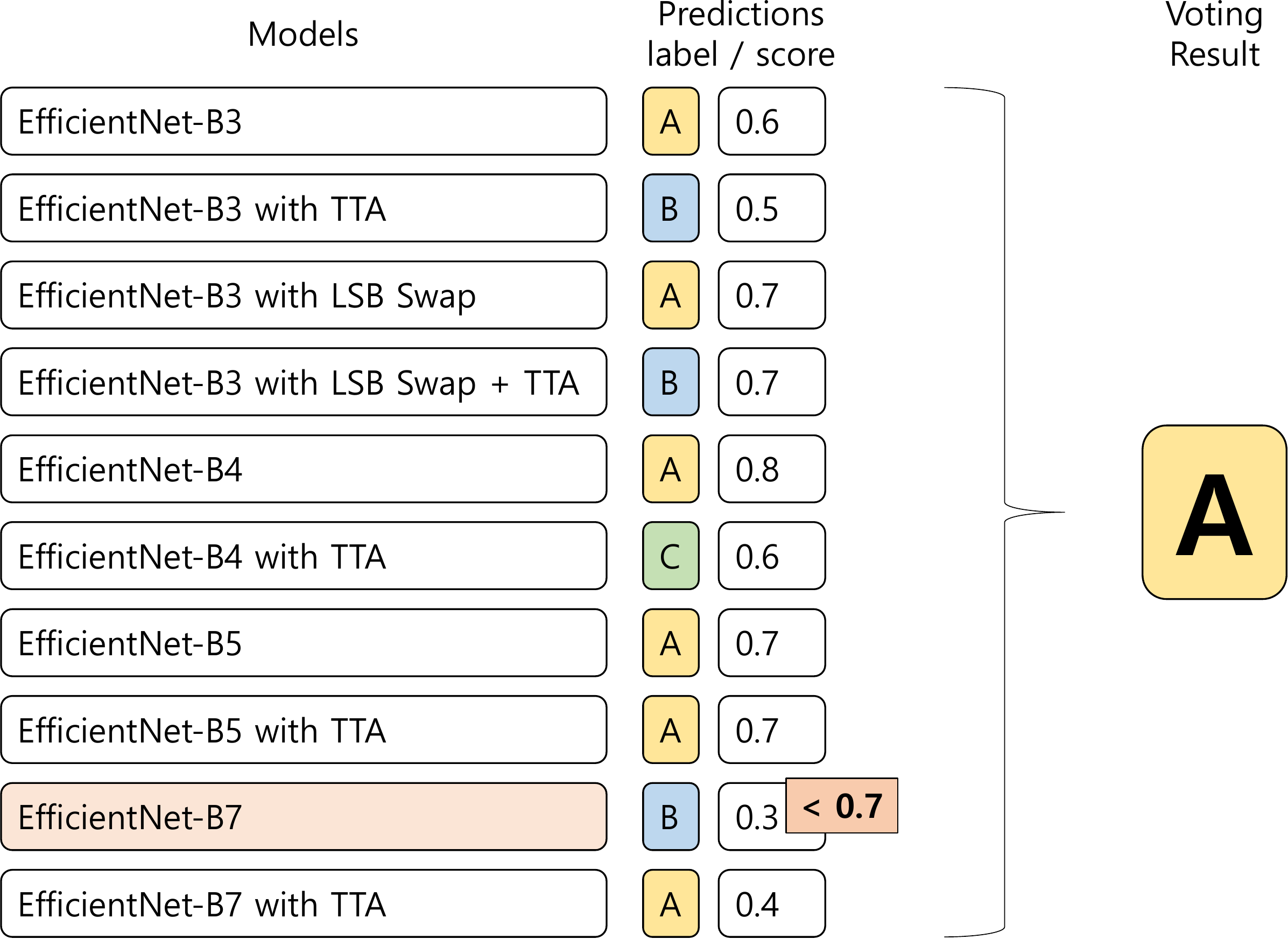}
\caption{An example of applying plurality voting ensemble to 10 models. Base model $B$ is set to EfficientNet-B7(orange box) and $threshold$ is set to 0.7.}
\label{fig:ensemble}
\end{figure}

Because there is no answer for the test data, accuracy could not be calculated. However, in the case of EfficientNet-B3, the accuracy could be obtained by submitting it to the competition(EfficientNet-B3 in Table~\ref{table:2}). Because the size of EfficientNet-B3 model is still small, we use other larger models as shown in Table~\ref{table:1}. Then, we applied plurality voting ensemble as shown in Figure~\ref{fig:ensemble} with 10 trained models using EfficientNet-B7 as the base model(Ensemble model in Table~\ref{table:2}). It has improved 0.09 accuracy more than the EfficientNet-B3.

\setlength{\tabcolsep}{4pt}
\begin{table}
\begin{center}
\caption{The accuracy we got from submitting to the competition.}
\label{table:2}
\begin{tabular}{ll}
\hline\noalign{\smallskip}
Model & Accuracy\\
\noalign{\smallskip}
\hline
\noalign{\smallskip}
EfficientNet-B3 & 0.5823\\
Ensemble model & 0.6730\\
\hline
\end{tabular}
\end{center}
\end{table}
\setlength{\tabcolsep}{1.4pt}

\section{Conclusions}
We apply techniques such as LSB Swap and Focal Cosine Loss to train with only a small amount of data. In addition, even if it is difficult to predict with the best single model, the final result can be supplemented through the proposed ensemble method.

Because the deadline is set for the competition, it is hard to test all the techniques. For example, LSB Swap data augmentation technique apply only to EfficientNet-B3. However, considering that the training loss of EfficientNet-B3 applied with LSB Swap is lower than the one not applied, we think it will be effective in other models as well. All source code used for training and testing is provided~\cite{kbjcode}, and we informs that pre-trained weights and external data are not used.

\clearpage
%
%
\bibliographystyle{splncs04}
\bibliography{egbib}
\end{document}